\pdfoutput=1

\documentclass[11pt]{article}

\usepackage[final]{acl}

\usepackage{times}
\usepackage{latexsym}
\usepackage{enumitem}
\usepackage{booktabs}
\usepackage{multirow}
\usepackage{subcaption}
\usepackage{graphicx}
\usepackage{amsfonts}
\usepackage{amsmath} 
\usepackage{tikz}
\usepackage{hyperref}
\usepackage{pgfplots}
\usetikzlibrary{
  positioning,
  shapes.geometric,
  shadows,
  arrows.meta,
  backgrounds,
  calc
}
\usepackage{xcolor}
\usepackage{tabularx}

\definecolor{bg}{RGB}{0,0,0}
\definecolor{gridcol}{RGB}{100,100,100}
\definecolor{textcol}{RGB}{200,200,200}
\definecolor{arrowcol}{RGB}{0,255,150}
\definecolor{selblue}{RGB}{60,120,220}
\definecolor{finalred}{RGB}{220,50,50}
\definecolor{finalborder}{RGB}{255,200,0}

\newcommand{\CircledA}[1]{%
    \tikz[baseline=(char.base)]{
        \node[shape=circle,draw,inner sep=2pt, color=blue, fill=green] (char) {#1};}
}
\newcommand{\CircledC}[1]{%
    \tikz[baseline=(char.base)]{
        \node[shape=circle,draw,inner sep=2pt, color=blue, fill=red] (char) {#1};}
}
\usepackage[T1]{fontenc}

\usepackage[utf8]{inputenc}

\usepackage{microtype}

\usepackage{inconsolata}

\usepackage{graphicx}
\usepackage{tcolorbox}

\usepackage[colorinlistoftodos,prependcaption,textsize=tiny]{todonotes}

\title{Date Fragments: A Hidden Bottleneck of Tokenisation \\for Temporal Reasoning}

\author{Gagan Bhatia\textsuperscript{1}\,
Maxime Peyrard\textsuperscript{2}\,
Wei Zhao\textsuperscript{1} \\[0.4em]
  \textsuperscript{1}University of Aberdeen \,
  \textsuperscript{2}Université Grenoble Alpes \& CNRS\\
  \texttt{\{g.bhatia.24,wei.zhao\}@abdn.ac.uk}
  }

\begin{document}
\maketitle
\begin{abstract}
Modern BPE tokenisers often split calendar dates into meaningless fragments, e.g., “20250312” $\rightarrow$ “202”, “503”, “12”, inflating token counts and obscuring the inherent structure needed for robust temporal reasoning. In this work, we (1) introduce a simple yet interpretable metric, termed date fragmentation ratio, that measures how faithfully a tokeniser preserves multi-digit date components; (2) release \textsc{DateAugBench}, a suite of 6500 examples spanning three temporal reasoning tasks: context-based date resolution, format-invariance puzzles, and date arithmetic across historical, contemporary, and future time periods; and (3) through layer-wise probing and causal attention-hop analyses, uncover an emergent date-abstraction mechanism whereby large language models stitch together the fragments of month, day, and year components for temporal reasoning. Our experiments show that excessive fragmentation correlates with accuracy drops of up to 10 points on uncommon dates like historical and futuristic dates. Further, we find that the larger the model, the faster the emergent date abstraction heals date fragments. Lastly, we observe a reasoning path that LLMs follow to assemble date fragments, typically differing from human interpretation (year $\rightarrow$ month $\rightarrow$ day). Our datasets and code are made publicly available \href{https://github.com/gagan3012/date-fragments}{here}.
\end{abstract}

\section{Introduction}

Understanding and manipulating dates is a deceptively complex challenge for modern large language models (LLMs). Unlike ordinary words, dates combine numeric and lexical elements in rigidly defined patterns—ranging from compact eight‐digit strings such as \texttt{20250314} to more verbose forms like “March 14, 2025” or locale‐specific variants such as “14/03/2025.” Yet despite their structured nature, these date expressions often fall prey to subword tokenisers that fragment them into semantically meaningless pieces. A tokeniser that splits “2025-03-14” into “20”, “25”, “-0”, “3”, “-1”, “4” not only inflates the token count but also severs the natural boundaries of year, month, and day. This fragmentation obscures temporal cues and introduces a hidden bottleneck: even state-of-the-art LLMs struggle to resolve, compare, or compute dates accurately when their internal representations have been so badly fragmented. This issue has a  critical impact on real-world applications:
\begin{figure}[t]
  \centering
  \includegraphics[width=\columnwidth]{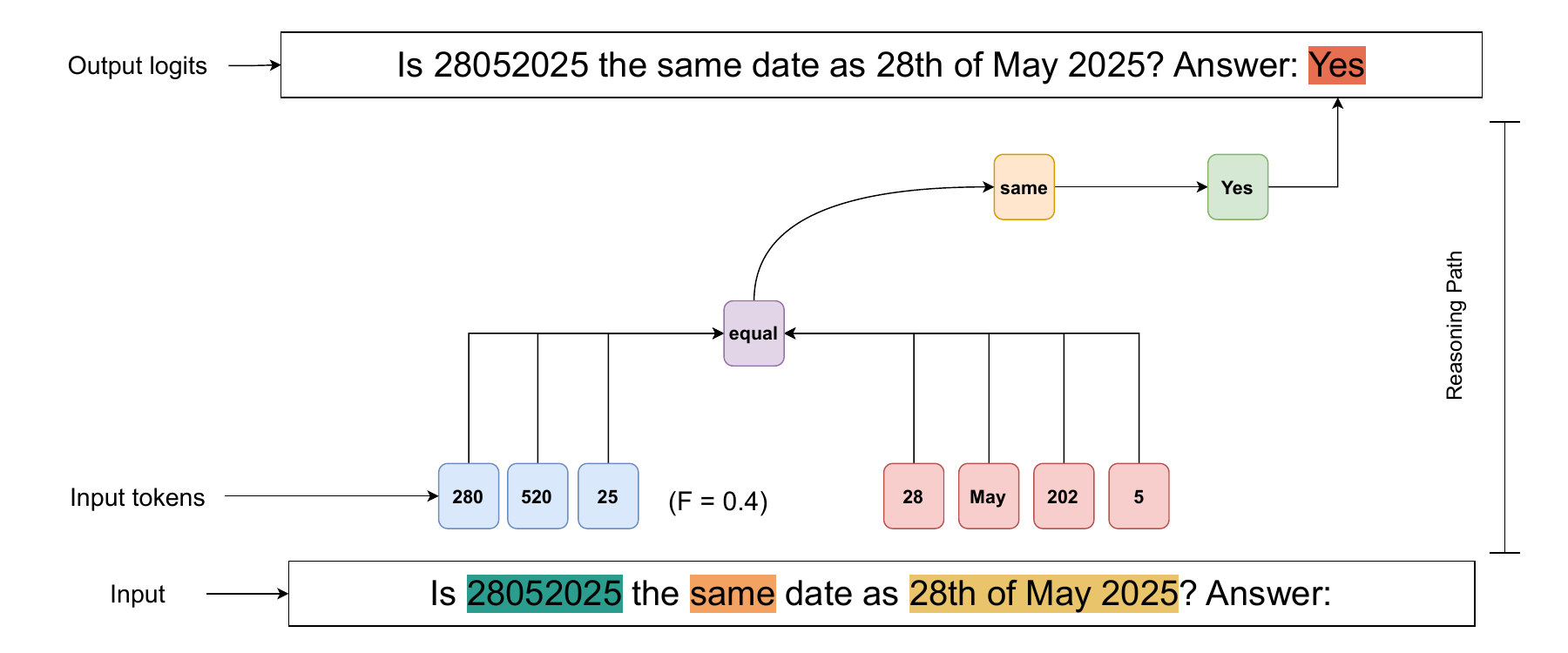}
  \caption{Internal processing of dates for temporal reasoning.
  Here F=0.4 shows the date fragmentation ratio.}
  \label{fig: mainexp}
\end{figure}

Mis‐tokenised dates can undermine scheduling and planning workflows, leading to erroneous calendar invites or appointments \cite{vasileiou2024synergistic}. They can skew forecasting models in domains ranging from time‐series analysis \cite{tan2024useful,chang2023llm4ts} to temporal knowledge graph reasoning \cite{wang2024dynamic}. In digital humanities and historical scholarship, incorrect splitting of date expressions may corrupt timelines and misguide interpretative analyses \cite{zeng2024histolens}. As LLMs are increasingly deployed in cross‐temporal applications, such as climate projection \cite{wang2024exploring}, economic forecasting \cite{carriero2024macroeconomic, bhatia2024fintralfamilygpt4level}, and automated curriculum scheduling \cite{vasileiou2024synergistic}, the brittleness introduced by subword fragmentation poses a risk of propagating temporal biases and inaccuracies into downstream scientific discoveries and decision‐making systems \cite{tan2024useful}.

In this work, we provide a pioneer outlook on the impact of date tokenisation on downstream temporal reasoning. Figure \ref{fig: mainexp} illustrates how dates are processed internally for temporal reasoning. Our contributions are summarized as follows:
\begin{enumerate}[label=(\roman*)]
\setlength\itemsep{-0.4em}
    \item We introduce \textsc{DateAugBench}, a benchmark dataset comprising 6,500 examples with 21 date formats. It is leveraged to evaluate a diverse array of LLMs from 8 model families in three temporal reasoning tasks.
    \item We present date fragmentation ratio, a metric that measures how fragmented the tokenisation outcome is compared to the actual year, month, and day components. We find that the fragmentation ratio generally correlates with temporal reasoning performance, namely that the more fragmented the tokenisation, the worse the reasoning performance.
    \item We analyse internal representations by tracing how LLMs “heal’’ a fragmented date embeddings in their layer stack---an emergent ability that we term \emph{date abstraction}. We find that larger models can quickly compensate for date fragmentation at early layers to achieve high accuracy for date equivalence reasoning.
    \item We leverage causal analysis to interpret how LLMs stitch date fragments for temporal reasoning. Our results show that LLMs follow a reasoning path that is typically not aligned with human interpretation (year $\rightarrow$ month $\rightarrow$ day), but relies on subword fragments that statistically represent year, month, and day, and stitch them in a flexible order that is subject to date formats.
\end{enumerate}

Our work fills the gap between
tokenisation research \cite{goldman2024unpacking,schmidt2024tokenization} and temporal reasoning \cite{su2024timo,fatemi2024test}, 
and we suggest future work to consider date-aware vocabularies and adaptive tokenisers to ensure that date components remain intact. 

\section{Related Works} \label{related_works}
\paragraph{Tokenisation as an information bottleneck.}  
Recent scholarship interrogates four complementary facets of sub-word segmentation: (i) \emph{tokenisation fidelity}, i.e.\ how closely a tokeniser preserves semantic units: Large empirical studies show that higher compression fidelity predicts better downstream accuracy in symbol-heavy domains such as code, maths and dates \cite{goldman2024unpacking,schmidt2024tokenization}; (ii) \emph{numeric segmentation strategies} that decide between digit-level or multi-digit units: Previous work demonstrates that the choice of radix-single digits versus 1-3 digit chunks induces stereotyped arithmetic errors and can even alter the complexity class of the computations LLMs can realise \cite{singh2024tokenization,zhou2024scaling};
(iii) \emph{probabilistic or learnable tokenisers} whose segmentations are optimised jointly with the language model: Theory frames tokenisation as a stochastic map whose invertibility controls whether maximum-likelihood estimators over tokens are consistent with the underlying word distribution \cite{gastaldi2024the,rajaraman2024toward} and (iv) \emph{pre-/post-tokenisation adaptations} that retrofit a model with a new vocabulary:
\citet{zheng2024adaptive} introduce an \emph{adaptive tokeniser} that co-evolves with the language model, while \citet{liu2025superbpe} push beyond the “sub-word” dogma with \emph{SuperBPE}, a curriculum that first learns subwords and then merges them into cross-whitespace “superwords”, cutting average sequence length by 27 \%.  
Complementary studies expose and correct systematic biases introduced by segmentation \cite{phan2024tokenbias} and propose \emph{trans-tokenisation} to transfer vocabularies across languages without re-training the model from scratch \cite{remy2024transtokenization}. Our work builds on these insights but zooms in on calendar dates---a hybrid of digits and lexical delimiters whose multi-digit fields are routinely shredded by standard BPE, obscuring cross-field regularities crucial for temporal reasoning.

\paragraph{Temporal reasoning in large language models.}  
Despite rapid progress on chain-of-thought and process-supervised reasoning, temporal cognition remains a conspicuous weakness of current LLMs.  Benchmarks such as \textsc{TimeBench} \cite{chu2024timebenchcomprehensiveevaluationtemporal}, \textsc{TempReason} \cite{tan2023towards}, \textsc{Test-of-Time} \cite{fatemi2024test}, \textsc{MenatQA} \cite{wei2023menatqa} and \textsc{TimeQA} \cite{chen2021datasetansweringtimesensitivequestions} reveal large gaps between model and human performance across ordering, arithmetic and co-temporal inference.  Recent modelling efforts attack the problem from multiple angles: temporal-graph abstractions \cite{xiong2024large}, instruction-tuned specialists such as \textsc{Timo} \cite{su2024timo}, pseudo-instruction augmentation for multi-hop QA \cite{tan2023towards}, and alignment techniques that re-ground pretrained models to specific calendar years \cite{zhao2024set}.  Yet these approaches assume a faithful internal representation of the input dates themselves. By introducing the notion of \emph{date fragmentation} and demonstrating that heavier fragmentation predicts up to ten-point accuracy drops on \textsc{DateAugBench}, we uncover a failure mode that is \emph{orthogonal} to reasoning algorithms or supervision: errors arise before the first transformer layer, at the level of subword segmentation.  Addressing this front-end bottleneck complements existing efforts to further improve LLMs for temporal reasoning.
\section{DateAugBench}\label{data}

\begin{table*}[ht]
\centering
\footnotesize
\resizebox{\textwidth}{!}{%
2\begin{tabularx}{\textwidth}{
  l             
  c             
  c             
  c             
  X             
  X             
}
\toprule
\multirow{2}{*}{\textbf{Dataset and Task}} 
  & \multirow{2}{*}{\textbf{\# Formats}} 
  & \multirow{2}{*}{\textbf{\# Raw}} 
  & \multirow{2}{*}{\textbf{Size}} 
  & \multicolumn{2}{c}{\textbf{Evaluation}} \\
\cmidrule(lr){5-6}
  &  &  &  & \textbf{Example} 
      & \textbf{GT} \\
\midrule
Context based              & 6  & 500  & 3000  & Which team did Omid Namazi play for in 06/10/1990?                   & Maryland Bays                                   \\\midrule
Date Format Switching    & 10 & 150  & 1500  & Are 20251403 and March 14th 2025 referring to the same date?         & Yes                                             \\\midrule
Date Arithmetic            & 5  & 400  & 2000  & What date is 10,000 days before 5/4/2025?                            & 18 November 1997; 17 December 1997              \\
\midrule
\textbf{Total}                       & 21 & 1500 & 6500  &  &  \\
\bottomrule
\end{tabularx}
}
\caption{Overview and examples of task splits in \textsc{DateAugBench}.}
\label{tab:date_datasets}
\end{table*}

We introduce \textsc{DateAugBench}, 
benchmark designed to isolate the impact of date tokenisation on temporal reasoning in LLMs. \textsc{DateAugBench} comprises 6,500 augmented examples drawn from two established sources, \textsc{TimeQA} \cite{chen2021datasetansweringtimesensitivequestions} and \textsc{TimeBench} \cite{chu2024timebenchcomprehensiveevaluationtemporal}, distributed across three tasks splits (see Table~\ref{tab:date_datasets}). Across all the splits, our chosen date formats cover a spectrum of common regional conventions (numeric with slashes, dashes, or dots; concatenated strings; two-digit versus four-digit years) and deliberately introduce fragmentation for atypical historical (e.g.\ “1799”) and future (e.g.\ “2121”) dates. This design enables controlled measurement of how tokenisation compression ratios and subsequent embedding recovery influence temporal reasoning performance.

\paragraph{Context-based task.} 
In the \emph{Context-based} split, we sample 500 question–context pairs from \textsc{TimeQA}, each requiring resolution of a date mentioned in the passage (e.g.\ Which team did Omid Namazi play for in 06/10/1990?). Every date expression is systematically rendered in six canonical serialisations---including variants such as \texttt{MM/DD/YYYY}, \texttt{DD-MM-YYYY}, \texttt{YYYY.MM.DD} and concatenations without delimiters---yielding 3,000 examples that jointly probe tokenisation fragmentation and contextual grounding.

\paragraph{Simple Format Switching task.}
The \emph{Simple Format Switching} set comprises 150 unique date pairs drawn from \textsc{TimeBench}, posed as binary same-day recognition questions (e.g.\ “Are 20251403 and 14th March 2025 referring to the same date?”). Each pair is presented in ten different representations, spanning slash-, dash-, and dot-delimited formats, both zero-padded and minimally notated, to stress-test format invariance under maximal tokenisation drift. This produces 1,500 targeted examples of pure format robustness. We also have examples where the dates are not equivalent, complicating the task.

\paragraph{Date Arithmetic task.}
The \emph{Date Arithmetic} split uses 400 arithmetic instances from \textsc{TimeBench} (e.g.\ What date is 10,000 days before 5/4/2025?). With the base date serialised in five distinct ways---from month-day-year and year-month-day with various delimiters to compact eight-digit forms. This results in 2,000 examples that examine the model’s ability to perform addition and subtraction of days, weeks, and months under various token fragmentation.  

\section{Experiment Design}\label{methodology}

\subsection{Date Tokenisation}

\paragraph{Tokenisers.}
For tokenisation analysis, we compare a deterministic, rule-based \emph{baseline tokeniser} against model-specific tokenisers. The baseline splits each date into its semantic components—year, month, day or Julian day—while preserving original delimiters. For neural models, we invoke either the OpenAI TikTok encodings (for \texttt{gpt-4}, \texttt{gpt-3.5-turbo}, \texttt{gpt-4o}, \texttt{text-davinci-003}) or Hugging Face tokenisers for open-source checkpoints. 
Every date string is processed to record the resulting sub-tokens, token count, and reconstructed substrings. 

\paragraph{Distance metric.}
To capture divergence from the ideal, we define a distance metric $\theta$ between a model’s token distribution and the baseline’s:
\begin{align}
\theta(\mathbf{t},\mathbf{b}) = 1 - \frac{\mathbf{t} \cdot \mathbf{b}}{|\mathbf{t}|,|\mathbf{b}|},
\end{align}
where $\mathbf{t}$ and $\mathbf{b}$ are vectors of sub-token counts for the model and baseline, respectively. A larger $\theta$ indicates greater sub-token divergence.

\paragraph{Date fragmentation ratio.}
Building on $\theta$, we introduce the \emph{date fragmentation ratio} $F$, which quantifies how fragmented a tokeniser’s output is relative to the baseline. We initialise $F=0.0$ for a perfectly aligned segmentation and apply downward adjustments according to observed discrepancies: a 0.10 penalty if the actual year/month/day components are fragmented (i.e., $\mathbf{1}_{\mathrm{split}}=1$)
, a 0.10 penalty if original delimiters are lost (i.e., $\mathbf{1}_{\mathrm{delimiter}}=1$), a 0.05 penalty multiplied by the token count difference $(N - N_b\bigr)$ between a tokeniser and the baseline,
and a $0.30\times\theta$ penalty for distributional divergence. The resulting $F\in[0,1]$ provides an interpretable score: values close to 0 denote minimal fragmentation, and values near 1 indicate severe fragmentation.

\begin{equation}
\begin{aligned}
F &= 0.10 \times \mathbf{1}_{\mathrm{split}}
    + 0.10 \times \mathbf{1}_{\mathrm{delimiter}} \\[6pt]
  &\quad
    + 0.05 \times \bigl(N - N_b\bigr)
    + 0.30 \times \theta
\end{aligned}
\label{eqn1}
\end{equation}

This date fragmentation ratio is pivotal because tokenisation inconsistencies directly impair a model’s ability to represent and reason over temporal inputs. When date strings are split non-intuitively, models encounter inflated token sequences and fragmented semantic cues, which can potentially lead to errors in tasks such as chronological comparison, date arithmetic, and context-based resolution. 

\paragraph{Validation of Date Fragmentation Ratio.}
To ensure our custom metric is well-founded, we performed a two-part validation. First, we conducted a human evaluation study, in which we found that our F metric's scores align strongly with human judgments of "fragmentation severity" (Spearman's $\rho = 0.84$), significantly outperforming standard metrics like BLEU ($\rho = 0.52$). Second, we used a data-driven approach to learn the metric's weights by training a model to predict the human severity scores from our fragmentation components. This process confirmed that our intuitively chosen weights accurately reflect the factors driving human perception of fragmentation. For a detailed breakdown of the human evaluation protocol and the data-driven validation, please see Appendix~\ref{app:metric_validation}.

\subsection{Temporal Reasoning Evaluation}
\paragraph{Models.}

We evaluate a spectrum of model ranging from 0.5 B to 14 B parameters: five open-source Qwen 2.5 models (0.5 B, 1.5 B, 3 B, 7 B, 14 B)~\cite{yang2024qwen2}, two Llama 3 models (3 B, 8 B)~\cite{touvron2023llama}, and two OLMo \cite{groeneveld2024olmo} models (1 B, 7 B). For comparison with state-of-the-art closed models, we also query the proprietary \texttt{GPT-4o} and \texttt{GPT-4o-mini} endpoints via the OpenAI API~\cite{openai2024gpt-4o}.

\paragraph{LLM-as-a-judge.}
To measure how date tokenisation affects downstream reasoning, we employ an LLM-as-judge framework using GPT-4o. For each test instance in \textsc{DateFragBench}, we construct a JSONL record that includes the question text, the model’s predicted answer, and a set of acceptable gold targets to capture all semantically equivalent date variants (e.g., both “03/04/2025” and “April 3, 2025” can appear in the gold label set). This record is submitted to GPT-4o via the OpenAI API with a system prompt instructing it to classify the prediction as \texttt{CORRECT}, \texttt{INCORRECT}, or \texttt{NOT ATTEMPTED}. A prediction is deemed \texttt{CORRECT} if it fully contains any one of the gold target variants without contradiction; \texttt{INCORRECT} if it contains factual errors relative to all gold variants; and \texttt{NOT ATTEMPTED} if it omits the required information. We validate GPT-4o’s reliability by randomly sampling 50 judged instances across all splits and obtaining independent annotations from four student evaluators. In 97\% of cases, GPT-4o’s judgments of model answers agree with the averaged human judgments across four student evaluators, with a Cohen’s $\kappa$ of 0.89 as the inter-annotator agreement, affirming the reliability of our automatic evaluation setup. 

\subsection{Internal Representations}
\begin{figure}[t]
\includegraphics[width=1\columnwidth]{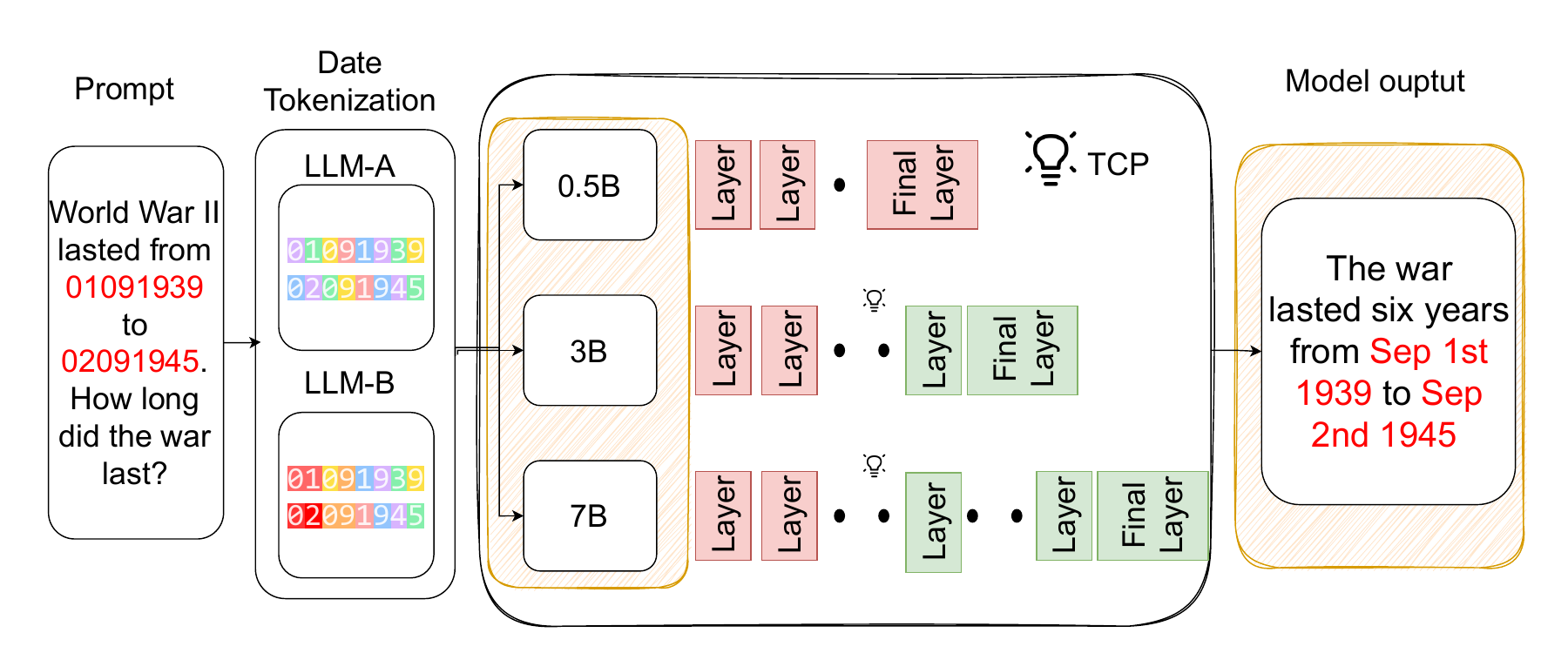}
\centering
\caption{Illustration of how LLMs with various model sizes process dates. TCP means Tokenization Compensation Point, defined as the first layer at which LLMs achieve above-chance accuracy
(see details in Sec. \ref{reasoning}).}
\label{fig: data_und}
\end{figure}
\paragraph{Layerwise probing.}
We use four Qwen2.5 \cite{yang2024qwen2} model checkpoints (0.5B, 1.5B, 3B, and 7B parameters) to trace how temporal information is processed internally across different layers. During inference, each question is prefixed with a fixed system prompt and a chain‐of‐thought cue, then passed through the model in evaluation mode. At each layer $i$, we extract the hidden‐state vector corresponding to the final token position, yielding an embedding $h_i \in \mathbb{R}^d$ for that layer. Repeating over all examples produces a collection of layer‐wise representations for positive and negative cases. We then quantify the emergence of temporal reasoning by training lightweight linear probes on these embeddings. For layer $i$, the probe is trained to distinguish “same‐date” (positive) vs “different‐date” (negative) examples. To explain when the model’s date understanding is achieved, we define the \emph{tokenisation compensation point} as the layer at which the model's representation correctly represents the date in the given prompt. We experiment with this idea across various model sizes, aiming to test our hypothesis: 
larger models would recover calendar‐level semantics from fragmented tokens at earlier stages, i.e., tokenisation compensation is accomplished at early layers, as illustrated in Figure \ref{fig: data_und}. 

\paragraph{Causal attention-hop analysis.}
\textcolor{black}{We introduce a framework intended to understand in which order date fragments are stitched together for LLMs to answer a temporal question. Figure \ref{fig: mainexp} depicts the idea of our framework: given an input prompt requiring a date resolution (e.g., “Is 28052025 the same date as 28th of May 2025?”), we define two sets of tokens: (1) \emph{concept tokens} corresponding to year, month, and day fragments, and (2) \emph{decision tokens} corresponding to the model answer (“yes” or “no”). Our framework aims to identify a stitching path for temporal reasoning, or reasoning path for short. A reasoning path is defined as a sequence of tokens containing date fragments and the model answer\footnote{The idea of reasoning paths was introduced by \citet{lindsey2025biology}, which we leverage to interpret how LLMs address date fragments for temporal reasoning.
}. Given that there are multiple potential paths, we score each path and select the highest‐scoring one as the LLM’s reasoning path for the given prompt. 
To score a reasoning path, our idea is the following: we identify when a date fragment or model answer is activated, by which input token and at which layer, and then determine how important each input token is for the date fragment and model answer. Our idea is implemented by using two different approaches: (i) next token prediction (\S\ref{sec:caha:tracing}): how likely a date fragment and model answer follows a given input token and (ii) token importance (\S\ref{sec:caha:intervention}): how important an input token is to a date fragment and model answer (by replacing the input token with a random token). Lastly, we combine the results of the two approaches to yield the final score of a reasoning path (\S\ref{sec:caha:scoring}).
This causal framework not only pinpoints \emph{where} and \emph{when} date fragments are activated, but also \emph{in which order} they are stitched together to yield the model answer.}

\section{Experiment Results}\label{results}
\subsection{Date fragmentation}

\begin{table}[ht]
\centering
\resizebox{\columnwidth}{!}{%
\begin{tabular}{lccccc}
\toprule
\textbf{Model}     & \textbf{Past} & \textbf{Near Past} & \textbf{Present} & \textbf{Future} & \textbf{Avg}   \\
\midrule
Baseline &0.00 &0.00 &0.00 &0.00 &0.00 \\
OLMo &0.15 &0.14 &0.07 &0.25 &0.15 \\
GPT-3 &0.17 &0.14 &0.06 &0.25 &0.16 \\
Llama 3 &0.29 &0.28 &0.27 &0.30 &0.29 \\
GPT-4o &0.32 &0.31 &0.22 &0.30 &0.29 \\
GPT-3.5 &0.47 &0.22 &0.26 &0.36 &0.33 \\
GPT-4 &0.36 &0.26 &0.29 &0.39 &0.33 \\
Qwen &0.58 &0.55 &0.49 &0.58 &0.55 \\
Gemma &0.58 &0.55 &0.49 &0.58 &0.55 \\
DeepSeek &0.58 &0.55 &0.49 &0.58 &0.55 \\
Llama 2 &0.63 &0.63 &0.63 &0.63 &0.63 \\
Phi &0.63 &0.63 &0.63 &0.63 &0.63 \\
\bottomrule
\end{tabular}}
\caption{Date fragmentation ratio across models and data splits over time. In case a family of model variants (Qwen, Gemma, DeepSeek and Phi) uses the same tokeniser, only the family name is referenced.}
\label{tab:temporal_Date_Fragmentation}
\end{table}

\begin{table}[!htp]
\centering
\resizebox{\columnwidth}{!}{%
\scriptsize
\begin{tabular}{lcccc}\toprule
\textbf{Models} &\textbf{Context Rlt} &\textbf{Fmt Switch} &\textbf{Date Arth.} &\textbf{Avg.} \\\midrule
GPT-4o-mini &\textbf{53.20} &95.66 &56.67 &\textbf{68.51} \\
OLMo-2-7B &32.13 &\textbf{97.24} &\textbf{64.72} &64.70 \\
Qwen2.5 14B &47.56 &94.56 &51.35 &64.49 \\
Qwen2.5 7B &39.56 &91.24 &40.56 &57.12 \\
Qwen2.5 3B &25.45 &90.10 &39.45 &51.67 \\
LLama3.1 8B &26.20 &90.22 &34.50 &50.31 \\
Qwen2.5 1.5B &21.32 &89.65 &32.34 &47.77 \\
Qwen2.5 0.5B &10.23 &88.95 &31.32 &43.50 \\
OLMo-2-1B &9.26 &90.09 &25.90 &41.75 \\
LLama3.2 3B &9.51 &88.45 &23.66 &40.54 \\
\bottomrule
\end{tabular}}
\caption{Average accuracies per task. Context Rlt stands for context based resolution, Fmt Switch refers to format switching, and Date Arth. refers to date arithmetic.}
\label{tab: res_all}
\end{table}

\paragraph{Cross-temporal performance.}
Table~\ref{tab:temporal_Date_Fragmentation}\
reports the mean date fragmentation ratio across four \textcolor{black}{time periods}---\emph{Past} (pre–2000), \emph{Near Past} (2000–2009), \emph{Present} (2010–2025), and \emph{Future} (post–2025)---for each evaluated model. A ratio of 0.00 signifies perfect alignment with our rule‐based baseline tokeniser, whereas higher values indicate progressively greater fragmentation.
The rule‐based \texttt{Baseline} unsurprisingly attains the maximal ratio of 0.00 in all periods, serving as a lower bound. Among neural architectures, OLMo \cite{groeneveld2024olmo} demonstrates the highest robustness, with an average fragmentation ratio of 0.15, closely followed by GPT-3 at 0.16. Both maintain strong fidelity across temporal splits, although performance dips modestly in the Future category (0.25), reflecting novel token sequences not seen during pre-training.

\begin{table}[ht]
  \centering
  \footnotesize
  \begin{tabular}{lcc}
    \toprule
    \textbf{Model} & \textbf{Tokenised output} & \textbf{Frag-ratio} \\ 
    \midrule
    Baseline  & \texttt{10 27 1606}                 & 0.00 \\
    OLMo      & \texttt{10 27 16 06}               & 0.34 \\
    Llama 3   & \texttt{102 716 06}                & 0.40 \\
    GPT-3     & \texttt{1027 16 06}                & 0.40 \\
    GPT-4o    & \texttt{102 716 06}                & 0.40 \\
    Gemma     & \texttt{1 0 2 7 1 6 0 6}           & 0.55 \\
    DeepSeek  & \texttt{1 0 2 7 1 6 0 6}           & 0.55 \\
    Cohere    & \texttt{1 0 2 7 1 6 0 6}           & 0.55 \\
    Qwen      & \texttt{1 0 2 7 1 6 0 6}           & 0.55 \\
    Phi 3.5   & \texttt{\_ 1 0 2 7 1 6 0 6}         & 0.60 \\
    Llama 2   & \texttt{\_ 1 0 2 7 1 6 0 6}         & 0.60 \\
    \bottomrule
  \end{tabular}
  \caption{Tokenisation of the MMDDYYYY string \texttt{``10271606''} across models.}
  \label{tab:mmddyyyy_tokenisation}
\end{table}

\paragraph{Impact of subtoken granularity.}
A closer look, from Table \ref{tab:mmddyyyy_tokenisation}, at sub-token granularity further explains these trends. Llama 3 \cite{touvron2023llama} and the GPT \cite{openai2023gpt-4} families typically segment each date component into three-digit sub-tokens (e.g., “202”, “504”, “03”), thus preserving the semantic unit of “MMDDYYYY” as compact pieces. OLMo \cite{groeneveld2024olmo} splits the date tokens into two digit tokens (e.g., “20”, “25”).
By contrast, Qwen \cite{yang2024qwen2} and Gemma \cite{gemmateam2024gemma2improvingopen} models break dates into single-digit tokens (e.g., “2”, “5”), whereas Phi \cite{abdin2024phi-3} 
divides it into single-digit tokens with an initial token (e.g.
“\_”, “2”, “0”, “2”, “5”), inflating the token count. Although single‐digit tokenisation can enhance models’ ability to perform arbitrary numeric manipulations (by treating each digit as an independent unit), it comes at the expense of temporal abstraction: the tight coupling between day, month, and year is lost, inflating the compression penalty and increasing the $\theta$ divergence from the baseline.

\subsection{\textsc{DateFragBench} Evaluation}

\begin{figure}[t]
  \centering
  \includegraphics[width=\linewidth]{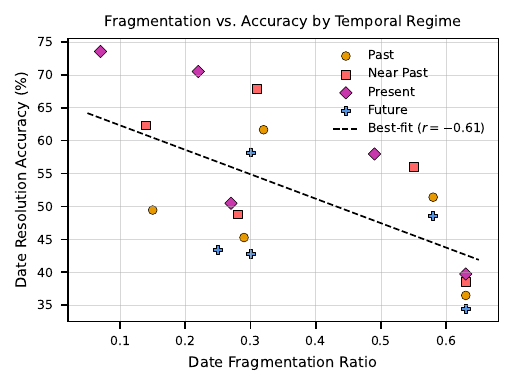}
  \caption{Date fragmentation ratio versus date resolution accuracy, stratified by four time periods and six LLMs: OLMo, Llama 3, GPT-4o, Qwen, Gemma, Phi.}
  \label{fig: res_ref}
\end{figure}

\begin{figure}[t]
  \centering
  \includegraphics[width=\columnwidth]{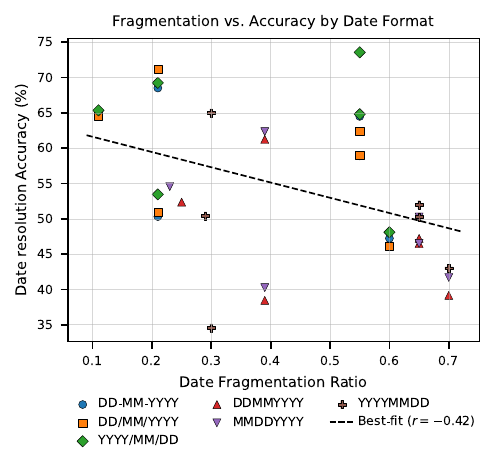}
  \caption{Date fragmentation ratio versus date resolution accuracy, stratified by six formats and six LLMs.}
  \label{fig:res_format}
\end{figure}

\paragraph{Performance on temporal reasoning tasks.}  
We compare model accuracies in three tasks: Context-based Resolution, Format Switching, and Date Arithmetic (see Table~\ref{tab: res_all}). All models effectively solve Format Switching (e.g.\,97.2\% for OLMo-2-7B, 95.7\% for GPT-4o-mini, 94.6\% for Qwen2.5-14B, 90.2\% for Llama3.1-8B). By contrast, Context Resolution and Arithmetic remain challenging: GPT-4o-mini scores 53.2\% and 56.7\%, Qwen2.5-14B 47.6\% and 51.4\%, Llama3.1-8B 26.2\% and 34.5\%, and OLMo-2-7B 32.1\% and 64.7\%, respectively. The fact that arithmetic performance consistently exceeds resolution suggests that, given a correctly tokenised date, performing addition or subtraction is somewhat easier than resolving the date within free text---which requires encyclopedic knowledge.

\paragraph{Correlating date fragmentation with model accuracy over time.}

Figure \ref{fig: res_ref} plots date fragmentation ratio against resolution accuracy, with 24 data points across six models and four temporal splits. Accuracy rises as we move from Past (1600-2000) to Near Past (2000–2009) and peaks in the Present (2010–2025), mirroring the negative correlation between fragmentation and accuracy (dashed line, Pearson correlation of $-0.61$). We note that the correlation is not particularly strong. This is because (i) for some models (e.g., Phi), their date fragmentation ratios remain unchanged across temporal data splits and (ii) models differ greatly by their sizes: a larger model could outperform a substantially smaller model in terms of temporal reasoning performance, even if the former has a much higher fragmentation ratio.

As seen from Table \ref{tab: res_ref}, GPT-4o-mini climbs from 61.7 \% in Past to 67.9 \% in Near Past, peaks at 70.5 \% for Present, and falls to 58.2 \% on Future dates. Qwen-2.5-14B and Llama-3.1-8B trace the same contour at lower absolute levels. OLMo-2-7B shows the steepest Near-Past jump (49.5 $\rightarrow$ 62.4 \%) and achieves the highest Present accuracy (73.6 \%), consistent with its finer-grained tokenisation of “20XX” patterns. These results indicate that while finer date tokenisation (i.e., lower fragmentation ratios) boosts performance up to contemporary references, today’s models still generalise poorly to genuinely novel (post-2025) dates, highlighting an open challenge for robust temporal reasoning.

\paragraph{Correlating date fragmentation with model accuracy over formats.}
Figure \ref{fig:res_format} plots model accuracy against date fragmentation ratio across six date formats and six LLMs. A moderate negative trend emerges (dashed line, Pearson correlation of $-0.42$): formats that contain explicit separators (DD-MM-YYYY, DD/MM/YYYY, YYYY/MM/DD) are tokenised into more pieces and, in turn, resolved more accurately than compact, separator-free strings (DDMMYYYY, MMDDYYYY, YYYYMMDD). 
As shown in Table \ref{tab:format_accuracy}, GPT-4o-mini tops every format and receives a moderate performance drop from 71.2 \% on DD/MM/YYYY to 61.2 \% on DDMMYYYY, with the highest overall average (66.3 \%). OLMo-2-7B and Qwen-2.5-14B both exceed 70 \% on the highly fragmented YYYY/MM/DD form, but slip into the low 50s on MMDDYYYY and YYYYMMDD. Lower date fragmentation ratio models, such as Llama-3.1-8B and Phi-3.5, lag behind; their accuracy plunges below 40 \%.
Even so, all models score much better on separator-rich formats compared to the date formats without separators. 
In summary, model accuracy is correlated to how cleanly a model can tokenise the string into interpretable tokens: more visual structure (slashes or dashes) means lower fragmentation, which suggests more straightforward reasoning, and in turn, leads to better performance.

\section{In which layer do LLMs compensate for  date fragmentation?}\label{reasoning}

\begin{figure*}[htbp]
  \centering
  \begin{subfigure}[b]{0.49\textwidth}
    \centering
    \includegraphics[width=\textwidth]{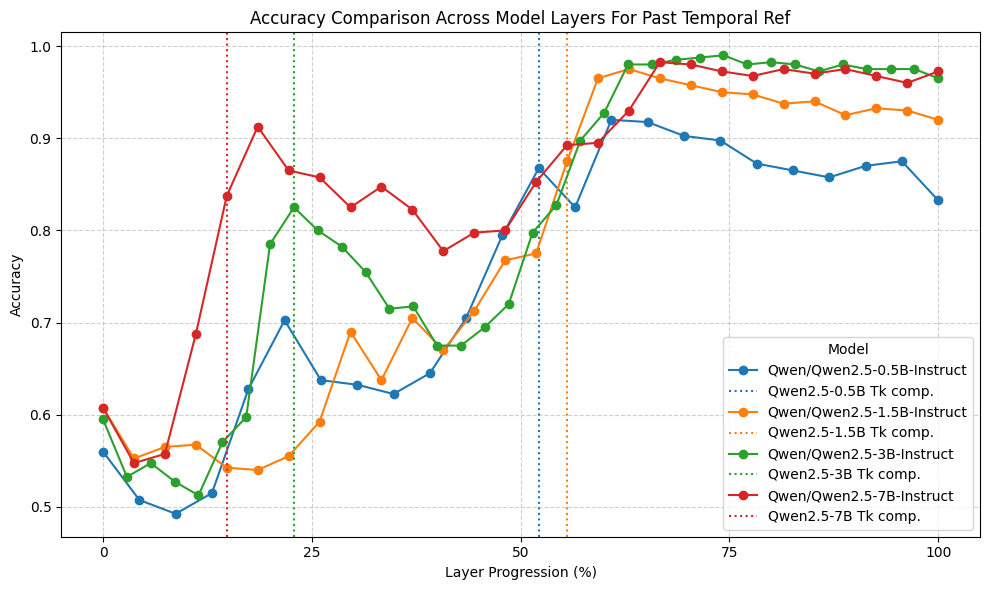}
    \caption{Past}
    \label{fig:past}
  \end{subfigure}
  \hfill
  \begin{subfigure}[b]{0.49\textwidth}
    \centering
    \includegraphics[width=\textwidth]{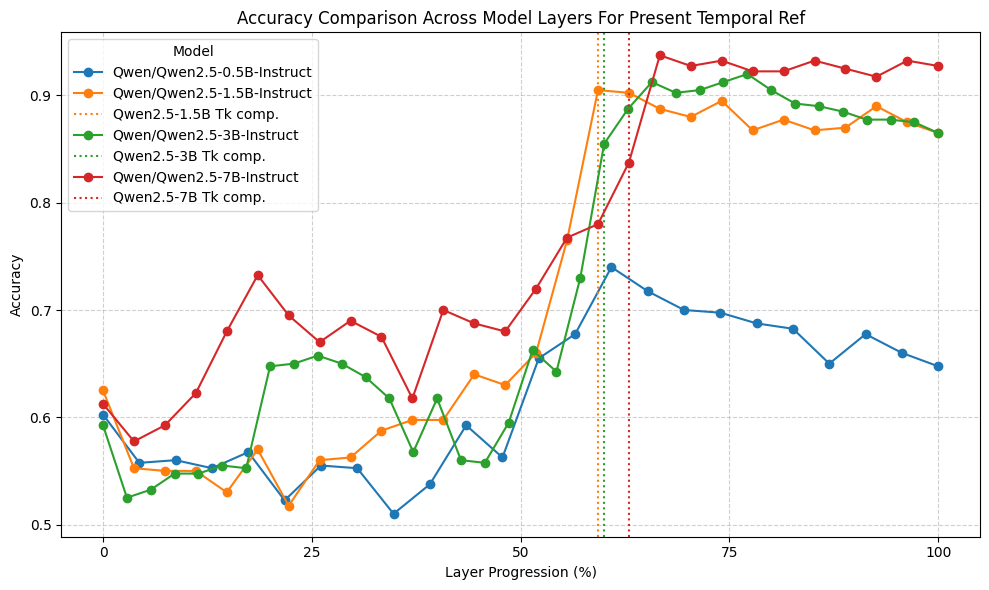}
    \caption{Present}
    \label{fig:present}
  \end{subfigure}
  \caption{Layer-wise accuracies in the two time periods: Past and Present.}
  \label{fig:temporal_comparison}
\end{figure*}


\begin{figure*}[t]
  \centering
  \includegraphics[width=0.85\linewidth]{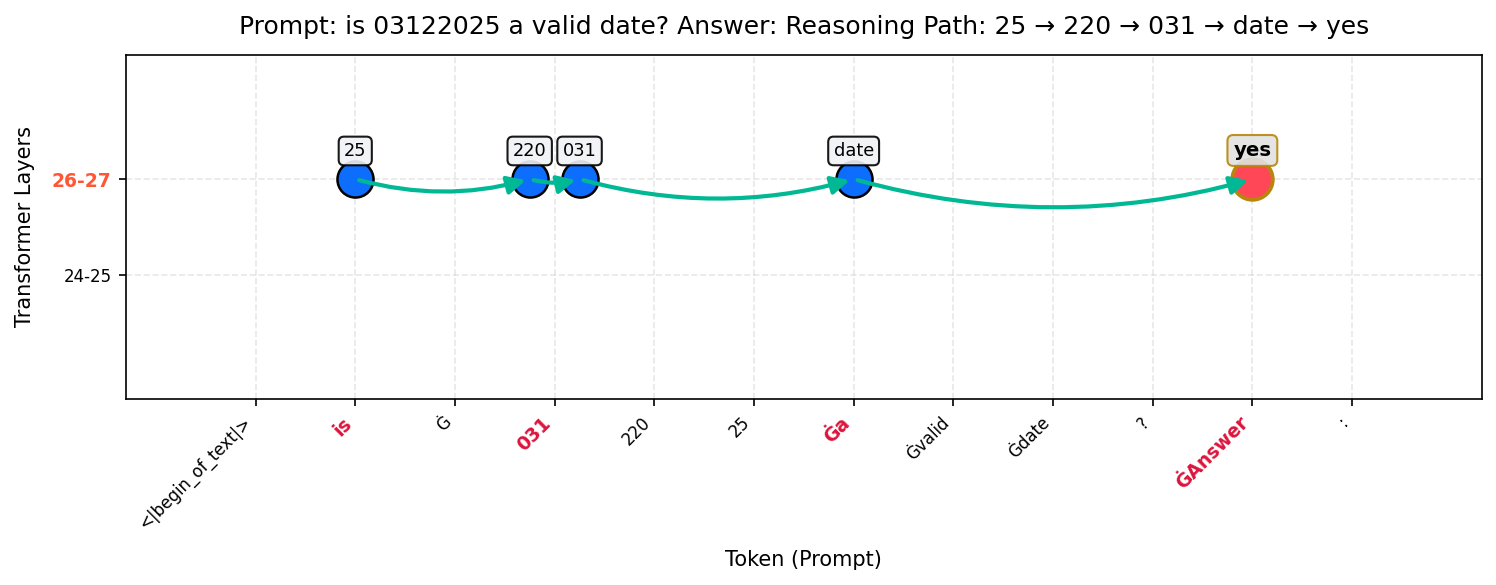}
  \caption{Reasoning path for the “03122025 is a valid date” prompt. 
  }
  \label{fig:date_flow}
\end{figure*}

\paragraph{Layerwise linear probing.}
To pinpoint in which layer
a model learns to recognize two equivalent dates, we define the \emph{tokenisation compensation point} (TCP) as the earliest layer at which a lightweight linear probe on the hidden state achieves above‐chance accuracy, which is defined as 80\%, on the date equivalence task.  Figure~\ref{fig:past} reports TCPs for the \textsc{Dates\_Past} benchmark (1600–2010): Qwen2.5-0.5B reaches TCP at layer 12 (50\% depth), Qwen2.5-1.5B at layer 15 (53.6\%), Qwen2.5-3B at layer 8 (22.2\%), and Qwen2.5-7B at layer 4 (14.3\%).  The leftward shift of the 3B and 7B curves suggests how larger models recover calendar‐level semantics from fragmented tokens more rapidly. Figure~\ref{fig:present} shows the \textsc{Dates\_Present} benchmark (2010–2025), where only the 1.5B, 3B, and 7B models surpass TCP—at layers 16 (57.1\%), 21 (58.3\%), and 17 (60.7\%), respectively---while the 0.5B model never does.  The deeper TCPs here reflect extra layers needed to recombine the two‐digit “20” prefix, which is fragmented unevenly by the tokeniser. In Figure~\ref{fig:future}, we evaluate \textsc{Dates\_Future} (2025–2599), where novel four‐digit sequences exacerbate fragmentation.  Remarkably, TCPs mirror the Past regime: layers 12, 15, 8, and 4 for the 0.5B, 1.5B, 3B, and 7B models, respectively.  This parallelism indicates that model scale dictates how quickly LLMs can compensate for date fragmentation to achieve high accuracy, even when dates are novel.

\paragraph{Tokenisation compensation point.}
Overall, we observe a sharp decline in TCP as model size increases: small models defer date reconstruction to middle layers, whereas the largest model does so within the first quarter of layers. Across all the three temporal benchmarks, TCP shifts steadily toward the first layers as model size grows. 

\section{How do LLMs stitch date fragments for temporal reasoning?}\label{reasoningv2}

\paragraph{Causal path tracing.}
To investigate how LLMs like Llama 3 \cite{touvron2023llama} internally stitch date fragments to yield a model answer, we apply our casual framework to identify the model's reasoning path over a specific prompt.
Figure \ref{fig:date_flow} plots model layers on the $y$ axis against prompt tokens (e.g., Is 03122025 a valid date?) on the $x$ axis. Green arrows mark the reasoning path with the highest score that is responsible for generating the answer “yes”. Date fragments “25”, “220”, “031”, and the model answer “yes” are activated in sequence at layer 26-27 by the input tokens “is”, “031”, “a” and “Answer” respectively.
 As such, the model 
performs a kind of discrete, step‐by‐step token aggregation, stitching together substrings of the input until a binary valid/invalid verdict emerges.

\paragraph{Misalignment between LLMs and human.}
In contrast, human readers parse dates by immediately mapping each component to a coherent temporal schema: “03” is March, “12” is day of month, “2025” is year, and then checking whether the day falls within the calendar bounds of that month. Humans bring rich world knowledge of calendars and leap‐year rules to bear in parallel. However, LLMs exhibit no explicit calendar “module”; instead, they rely on learned statistical associations between digit‐patterns and the training‐time supervisory signal for “valid date”. The reasoning path in Figure~\ref{fig:date_flow} thus illustrates a fundamentally different mechanism of date comprehension in LLMs, based on 
date fragments re‐routing rather than holistic semantic interpretation. We repeated causal tracing on 100 date strings in 6 different date formats to test whether the reasoning path difference between human and LLMs is consistent across date formats.
In most of cases, we observe that model reasoning paths are not aligned with human interpretation (year $\rightarrow$ month $\rightarrow$ day), rather rely on \emph{sub-word fragments} that statistically represent year, month, and day, and stitch these date fragments in a flexible order that is subject to date formats (see examples in Figures \ref{fig:date_flow2}-\ref{fig:date_flow3}). However, such a reasoning path becomes tricky when a date is greatly fragmented: given the date abstraction is learned from frequency rather than hard-coded rules, the abstraction is biased toward standard Western formats and contemporary years. As a result, a model often addresses popular dates (in the same format) with similar reasoning paths. However, the reasoning path becomes obscure
on rare, historical, or locale-specific strings outside the  distribution of pre-training data (see Figure \ref{fig:date_flow1}).

\section{Discussion} 
The moderate Pearson correlations of -0.61 (by temporal regime) and -0.42 (by date format) are a significant finding in themselves. They confirm that date fragmentation is a consistent and independent bottleneck, while also highlighting that it is not the sole factor influencing performance. The remaining variance is naturally explained by confounding factors such as model architecture, scale, and pre-training data exposure. For instance, a larger model may have a greater capacity to "heal" a poorly tokenised date, partially masking the negative impact of a high fragmentation ratio. Nonetheless, our results demonstrate that, all else being equal, higher fragmentation consistently predicts a drop in accuracy. This reveals a fundamental impediment that exists at the input level, before the model's core reasoning layers are even engaged.  

Our findings also shed light on the debate between memorisation and actual logical reasoning in LLMs. The performance disparity between "Present" dates and "Past" or "Future" dates (Figure \ref{fig: res_ref}) suggests that models heavily rely on statistical patterns and memorised facts from their pre-training data. For common contemporary dates, strong learned associations allow models to effectively parse and reason about them, even if tokenisation is suboptimal. However, for less frequent historical or novel future dates, this reliance on memorisation becomes a liability. The "date abstraction" mechanism struggles, and models must generalise from sparser data, leading to the observed accuracy drops. This contrasts with human date parsing, which leverages explicit, rule-based calendar knowledge rather than frequency-based recall.

\section{Conclusion} \label{conclusion}
In this paper, we identified date tokenisation as a critical yet overlooked bottleneck in temporal reasoning with LLMs. We demonstrated a correlation between date fragmentation and task performance in temporal reasoning, i.e., the more fragmented the tokenisation, the worse the reasoning performance.
Our layerwise and causal analyses in LLMs further revealed an emergent “date abstraction” mechanism that explains when and how LLMs understand and interpret dates. Our results showed that larger models can compensate for date fragmentation at early layers by stitching fragments for temporal reasoning, while the stitching process appears to follow a reasoning path that connects date fragments in a flexible order, differing from human interpretation from year to month to day.

\section*{Limitations}
While our work demonstrates the impact of date tokenisation on LLMs for temporal reasoning, there are several limitations. First, \textsc{DateAugBench} focuses on a finite set of canonical date serialisations and does not capture the full diversity of natural-language expressions (e.g., “the first Monday of May 2025”) or noisy real-world inputs like OCR outputs. Second, our experiments evaluate a representative but limited pool of tokenisers and model checkpoints (up to 14B parameters); therefore, the generalizability of date fragmentation ratio and our probing and causal analyses to very large models with 15B+ parameters remains unknown. Third, while the fragmentation ratio measures front-end segmentation fidelity, it does not account for deeper world-knowledge factors such as leap-year rules, timezone conversions, and culturally grounded calendar systems, all of which may influence temporal interpretation; further, the fragmentation ratio metric, though straightforward and interpretable, is not rigorously evaluated. Our work and the \textsc{DateAugBench} benchmark deliberately focus on a specific, foundational challenge: the tokenisation of explicit, multi-digit date strings in Anglo-centric Gregorian formats. This narrow scope allows us to isolate the impact of subword fragmentation on core temporal reasoning. However, this focus means we do not address the full complexity of temporal understanding, such as dates expressed in natural language (e.g., "the first Monday of May 2025"), dates with missing components, non-Gregorian calendars (e.g., Hijri, Hebrew), or dates represented with non-Latin numeral systems. We consider the extension to these diverse and important cases as critical future work that can build upon our foundational analysis of fragmentation.
Lastly, the core idea of our causal framework is 
inspired by \citet{lindsey2025biology}; however, our extension to temporal reasoning is not evaluated. Future work should extend to more diverse date expressions, broader model and tokeniser families, equipping tokenisers with external calendar-wise knowledge to improve further robust temporal reasoning, and conducting rigorous evaluation of the fragmentation ratio metric and the causal framework.

\section*{Ethical Considerations}

\textsc{DateAugBench} is derived solely from the public, research-licensed \textsc{TimeQA} and \textsc{TimeBench} corpora that do not contain sensitive data; our augmentation pipeline rewrites only date strings.  
However, our dataset focuses on 21 Anglo-centric Gregorian formats. Therefore, our data potentially reinforce a Western default and overlook calendars or numeral systems used in many other cultures, and our date fragmentation metric may over-penalise tokenisers optimised for non-Latin digits. 

\section*{Acknowledgements}
We thank the anonymous reviewers for their
thoughtful comments that greatly improved the work. We gratefully thank Madiha Kazi, Cristina Mahanta, and MingZe Tang for their support of conducting human evaluation for LLMs-as-judge. We also thank Ahmad Isa Muhammad for participating in early discussions.

\bibliography{custom}

\appendix
\section{Appendix}
\label{sec:appendix}

\subsection{Experiment Design}\label{sec:exp_app}

\paragraph{Implementation details of evaluation.}
The evaluation pipeline is implemented in Python and supports asynchronous API requests with retry logic, as well as multiprocessing to handle thousands of examples efficiently. After collecting GPT-4o’s label for each instance, we map \texttt{CORRECT}/\texttt{INCORRECT} \texttt{NOT ATTEMPTED} to categorical scores A, B, and C. We then compute three core metrics: overall accuracy (proportion of A scores), given-attempted accuracy (A over A+B), and the F1 score, defined as the harmonic mean of overall and given-attempted accuracy. Results are reported both globally and stratified by task split (Context-based, Format Switching, Date Arithmetic) and by temporal category (Past, Near Past, Present, Future). We adopt the sample prompts introduced in SimpleQA \cite{wei2024measuringshortformfactualitylarge} as our LLM-as-judge queries, ensuring consistent scoring instructions across all evaluations. Our specific prompt used for evaluation can be found in Table \ref{tab:eval_prompt}. We have presented our examples of LLM as a judge and human evaluation in Table \ref{tab:human_eval}.

\paragraph{Date ambiguities.}
We explicitly enumerate all valid variants in the gold label set for each example to handle multiple correct answers arising from date-format ambiguities. This ensures that any prediction matching one of these variants is marked correct, avoiding penalisation for format differences.

\paragraph{Synthetic benchmark construction for linear probing.}
We construct a suite of synthetic true–false benchmarks to isolate temporal reasoning across different reference frames. For the \textsc{Dates\_Past}, \textsc{Dates\_Present}, and \textsc{Dates\_Future} datasets, we sample 1,000 date–date pairs each, drawing calendar dates uniformly from the appropriate range and rendering them in two randomly chosen, distinct formatting patterns (\texttt{Ymd} vs \texttt{d/m/Y}). Exactly half of each set are “YES” examples (identical dates under different formats), which are our positive examples, and half are “NO” (different dates), which are our negative examples. All three datasets are balanced, shuffled, and split into equal positive and negative subsets to ensure fair probing.

\subsection{Causal Attention–Hop Analysis}\label{sec:caha}
\subsubsection{Next Token Prediction}
\label{sec:caha:tracing}

We treat each token in the prompt as a candidate “concept” to follow.  After the model processes the input, it produces a hidden vector $h_{\ell,p}$ \textcolor{black}{per token at position $p$ and layer $\ell$ }.  To see how \textcolor{black}{likely a concept $c$ (e.g., a date fragment and model answer) follows each input token,
we project $h_{\ell,p}$ through $W_{U}$ to yield the “probability” distribution of vocabulary tokens, and denote $s^{c}_{\ell,p}$ as the “probability” of the concept being the next token.}
\begin{equation}
  z_{\ell,p} \;=\; W_{U}\,h_{\ell,p},
  \qquad
  s^{c}_{\ell,p} \;=\; z_{\ell,p}[t_{c}],
  \label{eq:logit}
  \tag{1}
\end{equation}
where 
$t_c$ is the index of concept $c$ \textcolor{black}{in the vocabulary}. 

\subsubsection{Token Importance}
\label{sec:caha:intervention}

\textcolor{black}{To measure how important an input token is to a concept (e.g., a date fragment and model answer), }
we replace the token with an unrelated one (e.g., “Dallas” $\rightarrow$ “Chicago”) and compute \textcolor{black}{the probability drop of the concept incurred by the replacement, denoted as $I_{c,p}$ (which we compute only at the last layer)}: 
\begin{equation}
  I_{c,p}
  \;=\;
  \sigma\bigl(z_{p}[t_{c}]\bigr)
  \;-
  \;\sigma\bigl(\tilde{z}_p[t_{c}]\bigr),
  \label{eq:impact}
  \tag{2}
\end{equation}
where $\sigma$ is a softmax function. \textcolor{black}{The bigger the $I_{c,p}$, the more important the original token at position $p$ for the concept $c$}.

\subsubsection{Path scoring}
\label{sec:caha:scoring}

A \emph{reasoning path} $\mathcal{P}=(c_{1},\dots,c_{k})$ is \textcolor{black}{a sequence of tokens, indicating in which order date fragments are stitched together for LLMs
to answer a temporal question.}
We score each potential path by blending five components (ordering, activation strength, causal strength, gap penalty, 
and confidence in the final concept), into a single score:

\begin{align*}
  S(\mathcal{P})
  &=\;
      \alpha \times S_{\text{order}}
    + \beta  \times S_{\text{act}}
    + \gamma \times S_{\text{causal}}  \\
  &\quad
    - \eta \times S_{\text{gap}}
    + \kappa \times S_{\text{final}}
  \label{eq:score}
  \tag{1}
\end{align*}

Each term is designed to reward a different desirable property:

\begin{itemize}
  \item \textbf{Ordering}: we give points if the concepts appear in roughly left‐to‐right order in the prompt, and secondarily in increasing layer order:
  \begin{align*}
    S_{\text{order}}
    = 0.7\times \mathbf{1}[p_{1}\le\cdots\le p_{k}] \\
    + 0.3 \times \mathbf{1}[\ell_{1}\le\cdots\le\ell_{k}],
    \tag{2}
  \end{align*}
  where \textcolor{black}{$\mathbf{1}$ is an indicator function, $p_i=\max_{\ell,p}s^{c_i}_{\ell,p}$ indicating the position of the most important input token for a concept $c_i$ at the last layer. Similarly, $\ell_i$ is the layer at which an input token pays the most attention to the concept $c_i$.}

  \item \textbf{Activation}: \textcolor{black}{we compute the average position of the most important input token for a concept from 1 to $k$,} and 
  normalize by a threshold $\tau=0.2$, and clip to 1:
  \[
    S_{\text{act}}
    = \min\!\bigl(\tfrac{1}{k}\sum_{i=1}^{k} p_i\,/\,\tau,\;1\bigr),
    \tag{3}
  \]

  \item \textbf{Causal strength}: we use the \textcolor{black}{token importance score, denoted as} $d_i=|I_{c_{i+1},p_i}|$ between two adjacent concepts $c_{i+1}$ and $c_i$, upweight latter \textcolor{black}{scores}, and downweight missing links by a coverage term $\rho$, which is defined as the fraction of actual causal connections observed between consecutive concepts out of the total possible consecutive pairs in the path. The combined score then multiplies the weighted average of the \(d_i\) by \(\tfrac12 + \tfrac12\rho\), giving:
  \[
    S_{\text{causal}}
    =\bigl(\tfrac{\sum_i w_i d_i}{\sum_i w_i}\bigr)
     \bigl(0.5+0.5\rho\bigr),
    \tag{4}
  \]
  where $w_i=0.5+0.5\frac{i-1}{k-2}$.

  \item \textbf{Gap penalty}: to discourage large jumps in position, we compute the mean gap $\bar g$ and apply a small multiplier $\lambda=0.1$:
  \[
    S_{\text{gap}} = 1 - \lambda\,\bar g,
    \quad S_{\text{gap}}\le1.
    \tag{6}
  \]
  This is done to encourage model paths to think step by step instead of directly jumping to the conclusion (yes/no).

  \item \textbf{Final confidence}: \textcolor{black}{We compute the position of the most important input token for the last concept $c_k$}:
  \[
    S_{\text{final}} = \max_{\ell,p}s^{c_k}_{\ell,p}.
    \tag{7}
  \]
\end{itemize}

The reasoning path with the highest total score $S(\mathcal{P})$ is chosen as the model’s reasoning path over a specific prompt. \textcolor{black}{We note that \emph{Ordering}, \emph{Activation}, \emph{Gap penalty} and \emph{Final confidence} components are built upon next token prediction signals \(s^{c}_{\ell,p}\), 
whereas the \emph{Causal strength} component is derived solely from token importance 
score
\(
I_{c_{i+1},\pi_i},
\)
i.e.\ the drop in the softmax probability for concept \(c_{i+1}\) when the token at position \(p_i\) is replaced.}

\subsection{Detailed Validation of the Date Fragmentation Ratio}
\label{app:metric_validation}

This appendix provides a detailed account of the two-part validation process for our custom date fragmentation ratio (F), demonstrating its alignment with human intuition and its empirical soundness.

\subsubsection{Human Evaluation of Fragmentation Severity}
\label{app:human_eval}
This study was designed to confirm that our F metric captures what humans perceive as semantic disruption in tokenized dates more effectively than general-purpose text similarity metrics.

\paragraph{Methodology.} We recruited five computer science graduate students, who were familiar with NLP but blind to our hypotheses, to serve as annotators. We created a stimulus set of 100 tokenised date strings, stratified to represent a wide range of models, date formats, and fragmentation levels from our experiments. For each item, annotators were shown the original date and the list of sub-tokens, and asked to rate the \textbf{“fragmentation severity”} on a 5-point Likert scale, according to the following rubric with examples:
\begin{itemize}
    \item \textbf{1 (No Fragmentation):} Tokens perfectly preserve the semantic components.
    \textit{Example: `10-27-1606` $\rightarrow$ `['10', '-', '27', '-', '1606']`}
    
    \item \textbf{2 (Minor Fragmentation):} Mostly preserved, with minor, non-ideal splits.
    \textit{Example: `1606` $\rightarrow$ `['16', '06']`}
    
    \item \textbf{3 (Moderate Fragmentation):} Core components are broken, making the structure harder to discern. Delimiters might be lost or numbers oddly grouped.
    \textit{Example: `10271606` $\rightarrow$ `['102', '716', '06']`}
    
    \item \textbf{4 (High Fragmentation):} Date split into many small pieces (e.g., single digits), though the original characters are easily reassembled.
    \textit{Example: `1606` $\rightarrow$ `['1', '6', '0', '6']`}
    
    \item \textbf{5 (Severe Fragmentation):} Tokenization completely obscures the date's structure, often by adding non-numeric tokens or creating highly unintuitive groupings.
    \textit{Example: `1606` $\rightarrow$ `['\_', '1', '6', '0', '6']`}
\end{itemize}
The human judgments were highly reliable, with a Krippendorff's Alpha for inter-annotator agreement of $\alpha = 0.81$.

\paragraph{Results.} We computed the Spearman's rank correlation coefficient ($\rho$) between the average human rating for each item and the scores from our F metric, BLEU, and character-level Edit Distance. As shown in Table \ref{tab:human_eval_appendix}, our F metric demonstrated a strong correlation with human ratings, far exceeding the general-purpose metrics. In Table \ref{tab:human_vs_metric_example}, we present examples from the human validation study.

\begin{table*}[ht]
    \centering
    \footnotesize
    \begin{tabular}{lccc}
        \toprule
        \textbf{Model} & \textbf{Tokenised output} & \textbf{Frag-ratio} & \textbf{Avg. Human Severity Rating (1-5)} \\
        \midrule
        Baseline       & \texttt{10 27 1606}           & 0.00 & 1.0 \\
        OLMo           & \texttt{10 27 16 06}          & 0.34 & 2.0 \\
        Llama 3        & \texttt{102 716 06}           & 0.40 & 3.4 \\
        GPT-4o         & \texttt{102 716 06}           & 0.40 & 3.4 \\
        Cohere         & \texttt{1 0 2 7 1 6 0 6}      & 0.55 & 4.6 \\
        Phi 3.5        & \texttt{\_ 1 0 2 7 1 6 0 6}   & 0.60 & 5.0 \\
        Llama 2        & \texttt{\_ 1 0 2 7 1 6 0 6}   & 0.60 & 5.0 \\
        \bottomrule
    \end{tabular}
    \caption{An illustrative example showing the strong correlation between the calculated fragmentation ratio (Frag-ratio) and the average human-perceived severity rating for the tokenisation of the MMDDYYYY string \texttt{"10271606"}. Higher scores in both metrics indicate greater fragmentation.}
    \label{tab:human_vs_metric_example}
\end{table*}

\begin{table*}[h!]
    \centering
    \begin{tabular}{lc}
        \toprule
        \textbf{Metric} & \textbf{Correlation with Human Ratings ($\rho$)} \\
        \midrule
        \textbf{Date Fragmentation Ratio (F)} & \textbf{0.84} \\
        BLEU Score                            & 0.52          \\
        Character-Level Edit Distance         & 0.21          \\
        \bottomrule
    \end{tabular}
    \caption{Spearman Correlation ($\rho$) of Metrics with Human Judgments of Fragmentation Severity.}
    \label{tab:human_eval_appendix}
\end{table*}

This result confirms that our specialised metric is necessary and effective, as it successfully quantifies the semantic disruption that humans perceive but that generic text metrics fail to capture.

\subsubsection{Data-Driven Validation of Metric Coefficients}
\label{app:data_driven_validation}
This analysis provides an empirical grounding for the weights used in our F metric's formula. By learning the weights from data, we can validate that our initial, intuitive design aligns with a more formal, data-driven approach.

\paragraph{Formal Problem Formulation.}
To directly tune our metric to align with human perception, we frame the task as a linear regression problem where the goal is to predict the average human severity rating. This setup is more straightforward for validating the metric's components against human judgment.
\begin{itemize}
    \item The target variable is the \textbf{Average Human Severity Rating}, a continuous score from 1 to 5, as described in Appendix \ref{app:human_eval}.
    \item The feature vector $\mathbf{x} \in \mathbb{R}^4$ consists of the four fragmentation components: $\mathbf{x} = [1_{\text{split}}, 1_{\text{delimiter}}, (N - N_b), \theta]$.
    \item We aim to learn a weight vector $\mathbf{w}$ such that: \\
    $\text{Avg. Human Severity Rating} \approx \mathbf{w}^T \mathbf{x} + \text{intercept}$.
\end{itemize}
We used a non-negative linear regression model, as each fragmentation component is hypothesised to increase, not decrease, the perceived severity. Features were standardised before training to ensure the learned coefficients were comparable.

\paragraph{Results and Confirmation.}
After fitting the model to our human evaluation data, we obtained a set of empirically derived coefficients. We normalised these weights to sum to 1 to compare them with the relative importance implied by our original formula. As Table \ref{tab:learned_weights_appendix} shows, the weights learned by predicting human ratings are remarkably similar to the normalised version of our original, intuitively set weights.

\begin{table*}[h!]
    \centering
    \begin{tabular}{lcc}
        \toprule
        \textbf{Fragmentation Component} & \textbf{Original Intuitive Weight} & \textbf{Empirically Learned Weight} \\
        & \textbf{(Normalized)} & \textbf{(from Human Ratings)} \\
        \midrule
        $1_{\text{split}}$ (Component Split)    & $0.10 / 0.55 \approx \textbf{0.1818}$ & \textbf{0.2015} \\
        $1_{\text{delimiter}}$ (Delimiter Loss) & $0.10 / 0.55 \approx \textbf{0.1818}$ & \textbf{0.1932} \\
        $N - N_b$ (Token Difference)            & $0.05 / 0.55 \approx \textbf{0.0909}$ & \textbf{0.1053} \\
        $\theta$ (Distributional Divergence)    & $0.30 / 0.55 \approx \textbf{0.5455}$ & \textbf{0.5000} \\
        \bottomrule
    \end{tabular}
    \caption{Comparison of Original (Normalised) and Empirically Learned Weights for the F Metric, using human ratings as the target variable.}
    \label{tab:learned_weights_appendix}
\end{table*}

This result provides strong empirical validation of our F metric's design from an alternative perspective. It demonstrates that our initial weights, chosen based on semantic principles, accurately reflect not only the impact on model performance but also the factors that drive human perception of the severity of fragmentation. The relative importance of the components remains consistent: distributional divergence ($\theta$) is the most significant factor, followed by major structural breaks (splits and delimiter loss), and finally by token count inflation. This confirms the robustness and validity of our metric's formulation.

\begin{table}[!htp]
\centering
\resizebox{\columnwidth}{!}{%
\scriptsize
\begin{tabular}{lrrrrr}\toprule
\textbf{Models} &\textbf{Past} &\textbf{Near Past} &\textbf{Present} &\textbf{Future} \\\midrule
GPT-4o-mini &61.66 &67.93 &70.51 &58.23 \\
OLMo-2-7B &49.45 &62.35 &73.56 &43.45 \\
Qwen2.5 14B &58.97 &64.80 &67.22 &55.69 \\
Qwen2.5 7B &51.41 &55.98 &57.98 &48.55 \\
Qwen2.5 3B &46.50 &50.25 &51.98 &43.91 \\
LLama3.1 8B &45.28 &48.82 &50.48 &42.76 \\
Qwen2.5 1.5B &42.99 &46.16 &47.69 &40.60 \\
Qwen2.5 0.5B &39.15 &41.68 &43.00 &36.98 \\
OLMo-2-1B &36.07 &38.09 &40.49 &34.07 \\
LLama3.2 3B &36.48 &38.57 &39.74 &34.46 \\
\bottomrule
\end{tabular}}
\caption{Model accuracy on context-based resolution across four data splits over time.}
\label{tab: res_ref}
\end{table}

\begin{table*}
  \centering
  \footnotesize
  \resizebox{\textwidth}{!}{%
  \begin{tabular}{lccccccc}
    \toprule
    \textbf{Model} & \textbf{DD-MM-YYYY} & \textbf{DD/MM/YYYY} & \textbf{YYYY/MM/DD} & \textbf{DDMMYYYY} & \textbf{MMDDYYYY} & \textbf{YYYYMMDD} & \textbf{Avg.} \\ 
    \midrule
    OLMo      & 64.70 & 64.56 & 65.35 & 52.35 & 54.56 & 50.41 & 58.65 \\
    Llama 3   & 50.31 & 50.89 & 53.45 & 38.45 & 40.24 & 34.56 & 44.65 \\
    GPT-4o    & 68.51 & 71.23 & 69.24 & 61.23 & 62.34 & 64.98 & 66.25 \\
    Qwen      & 64.49 & 62.35 & 73.56 & 46.50 & 50.25 & 51.98 & 58.19 \\
    Gemma     & 58.90 & 58.97 & 64.80 & 47.22 & 46.50 & 50.25 & 54.44 \\
    Phi       & 47.23 & 46.07 & 48.09 & 39.15 & 41.68 & 43.00 & 44.20 \\
    \bottomrule
  \end{tabular}
  }
  \caption{Model accuracy on context-based resolution across date formats.}
  \label{tab:format_accuracy}
\end{table*}

\begin{figure*}[t]
  \centering
  \includegraphics[width=\linewidth]{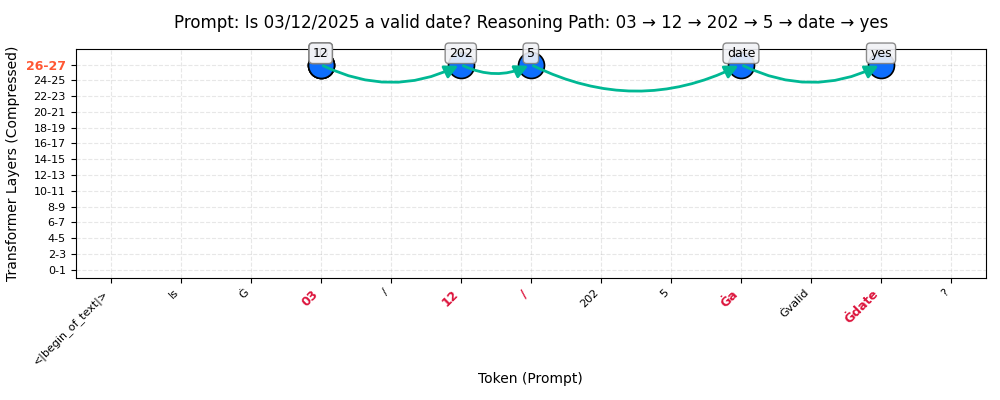}
  \caption{Reasoning path for the “03/12/2025 is a valid date” prompt.}
  \label{fig:date_flow2}
\end{figure*}

\begin{figure*}[t]
  \centering
  \includegraphics[width=\linewidth]{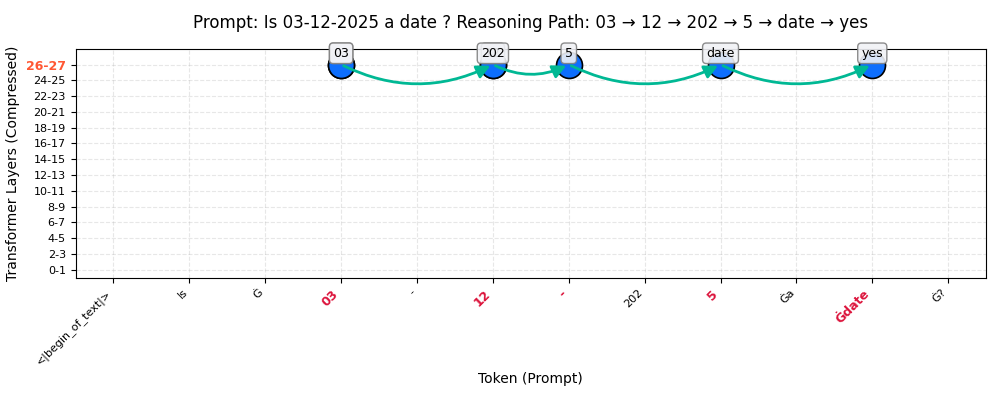}
  \caption{Reasoning path of the “03-12-2025 is a valid date” prompt. }
  \label{fig:date_flow3}
\end{figure*}

\begin{figure*}[t]
  \centering
  \includegraphics[width=\linewidth]{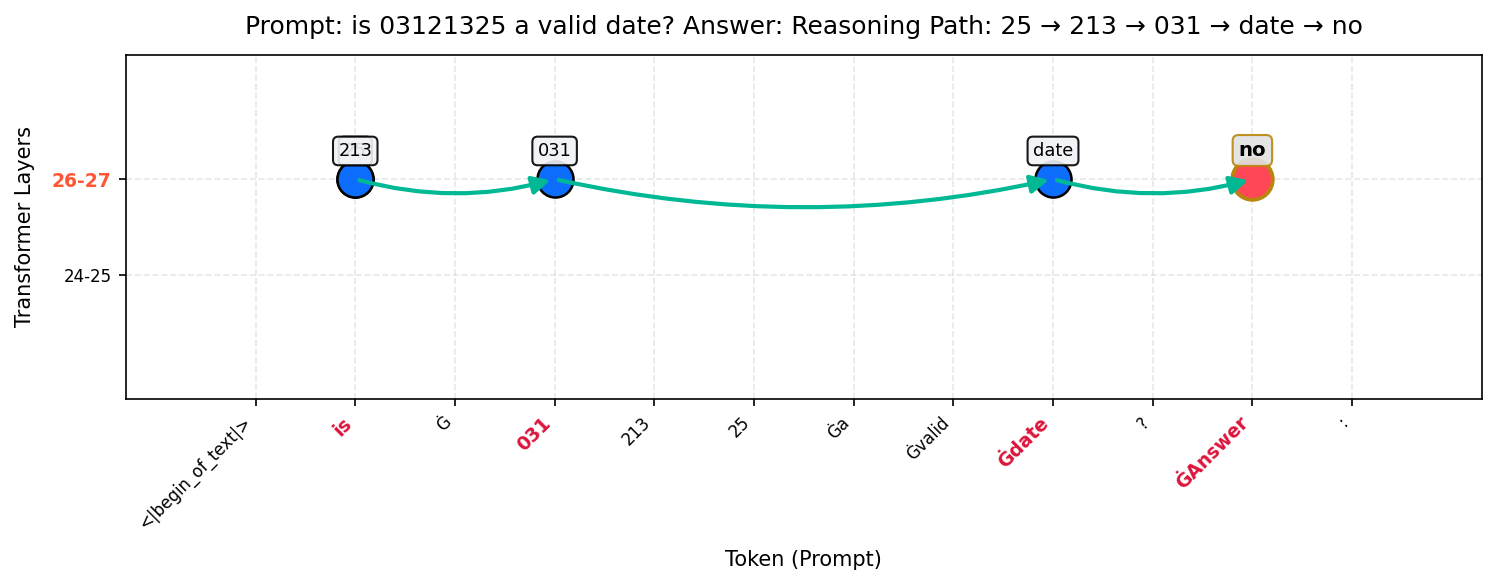}
  \caption{Reasoning path of the “03121325 is a valid date” prompt, where year $=1325$.}
  \label{fig:date_flow1}
\end{figure*}

\begin{figure*}[t]
  \centering
  \includegraphics[width=0.8\linewidth]{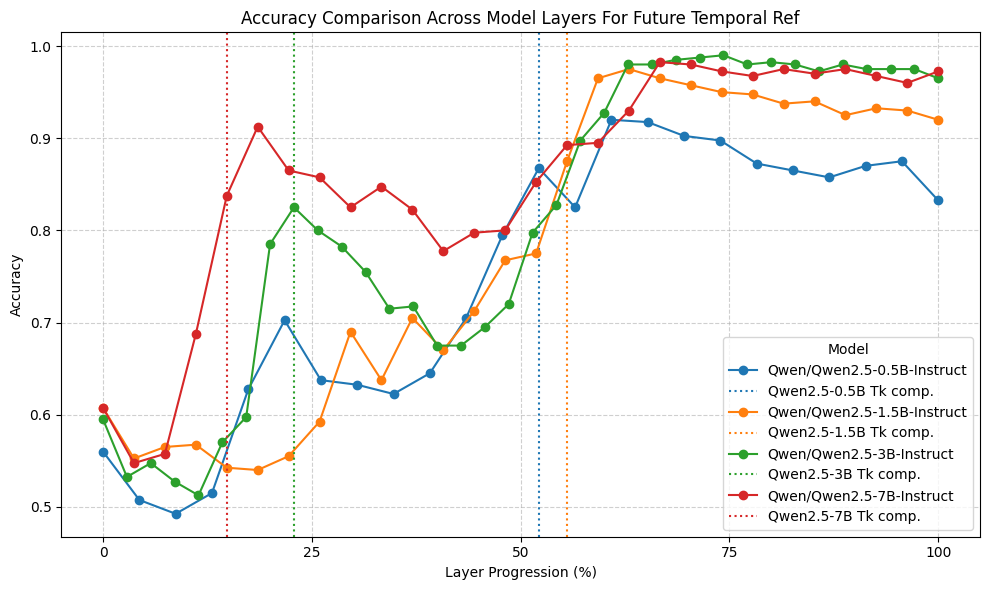}
  \caption{Layer-wise accuracies in the Future period.}
  \label{fig:future}
\end{figure*}

\begin{table*}[ht]
  \centering
  \begin{tcolorbox}[
      colback=blue!5!white,
      colframe=blue!75!black,
      title=LLM‑as‑Judge Evaluation Prompt]
\small
\textbf{Your task:} Evaluate one prediction at a time. You receive:
\begin{itemize}
  \item \textbf{Question} – the task prompt shown to the model
  \item \textbf{Gold target} – \emph{all} answers that are considered correct
  \item \textbf{Predicted answer} – the model’s response
\end{itemize}

Return \textbf{one letter only}:\\
\begin{tabular}{lll}
\textbf{A} & CORRECT        & prediction fully matches \emph{one} gold variant          \\
\textbf{B} & INCORRECT      & prediction contradicts or misses required info             \\
\textbf{C} & NOT\_ATTEMPTED & prediction refuses, guesses, or answers irrelevantly        \\
\end{tabular}
\\
[1.5pt]

\textbf{General rules}:
\begin{enumerate}\itemsep2pt
  \item Match semantics, ignore capitalisation, punctuation, order.
  \item If any statement contradicts the gold target, grade \textbf{B}.
  \item Hedging ("I think…") is fine if the correct info is present and no incorrect info is added.
  \item Partial answers are \textbf{B}. Typos that preserve meaning are allowed.
\end{enumerate}
\textbf{DateAugBench specifics}:\\[-6pt]
\begin{itemize}\itemsep2pt
  \item \textbf{Date format ambiguity}: gold lists every valid interpretation; accept any.
  \item \textbf{Date arithmetic}: prediction must match \emph{day, month, year} of a listed variant, any textual format allowed.
  \item \textbf{Format‑switch questions}: answer with any synonym of \texttt{Yes/True} or \texttt{No/False}.
  \item \textbf{Numeric answers} – must match the gold number to the last shown significant digit.
\end{itemize}

\textbf{Output format}\\
Return exactly one capital letter:
\[
\texttt{A}\quad\text{or}\quad\texttt{B}\quad\text{or}\quad\texttt{C}
\]
No additional text or punctuation.

\medskip
\textbf{Example template}
\begin{verbatim}
Question: {question}
Gold target: {target}
Predicted answer: {predicted_answer}
\end{verbatim}

\medskip
\textbf{Now grade:}
\[
\texttt{A}\quad\text{or}\quad\texttt{B}\quad\text{or}\quad\texttt{C}
\]
  \end{tcolorbox}
  \caption{LLM-as-Judge prompt used for comparing model and gold answers in the three DateAugBench tasks.}
  \label{tab:eval_prompt}
\end{table*}
\begin{table*}[ht]
    \centering
    \footnotesize
    \begin{tcolorbox}[
        colback=green!5!white,
        colframe=green!75!black,
        title=Human Evaluation]

        \begin{center}
            \Large \textbf{Context-based resolution}
        \end{center}
        \vspace{0.5em}

        \textbf{Prompt}: Who was the chair of Allgemeiner Deutscher Fahrrad-Club in 17/10/2016? \\ 
        \textbf{Gold Answer}: Ulrich Syberg \\ 
        \textbf{Model Prediction}: As of October 17, 2016, the Federal Chairman was Ulrich Syberg\\ 

        \textbf{Human Annotator Rating}: \; \CircledA{A} \hfill 
        \\[0.2em]
        \textbf{LLM-as-Judge Rating}: \; \CircledA{A} \hfill 

        \tcblower
        \begin{center}
            \Large \textbf{Date arithmetic}
        \end{center}
        \vspace{0.5em}

        \textbf{Prompt}: What date is 60 days after 05/01/1225? \\
        \textbf{Gold Answer}: March 6, 1225 , June 29, 1225\\
        \textbf{Model Prediction}: July 30, 1225\\

        \textbf{Human Annotator Rating}: \; \CircledC{B} \\
        \textbf{LLM-as-Judge Rating}: \; \CircledC{B} \\
    \end{tcolorbox}
    \caption{Human evaluation of LLM-as-judge.}
    \label{tab:human_eval}
\end{table*}

\end{document}